\documentclass[10pt,twocolumn,letterpaper]{article}

\usepackage{cvpr}              

\usepackage[dvipsnames]{xcolor}
\usepackage{times}
\usepackage{epsfig}
\usepackage{graphicx}
\usepackage{amsmath}
\usepackage{amssymb}
\usepackage{multirow}
\usepackage{makecell}
\usepackage{wrapfig}
\usepackage{dsfont}

\definecolor{cvprblue}{rgb}{0.21,0.49,0.74}
\usepackage[pagebackref,breaklinks,colorlinks,citecolor=cvprblue]{hyperref}
\usepackage[accsupp]{axessibility} 
\newcommand{\printfnsymbol}[1]{%
\textsuperscript{\@fnsymbol{#1}}%
}

\newcommand\blfootnote[1]{%
  \begingroup
  \renewcommand\thefootnote{}\footnote{#1}%
  \addtocounter{footnote}{-1}%
  \endgroup
}

\title{6D-Diff: A Keypoint Diffusion Framework for 6D Object Pose Estimation}

\author{Li Xu$^{1\dag}$
~~~ Haoxuan Qu$^{1\dag}$
~~~ Yujun Cai$^{2}$
~ Jun Liu$^{1\ddag}$ \\
\textsuperscript{1}Singapore University of Technology and Design ~~ \\
\textsuperscript{2}Nanyang Technological University \\
{\tt\small \{li\_xu,haoxuan\_qu\}@mymail.sutd.edu.sg, yujun001@e.ntu.edu.sg, }\\ 
{\tt\small  jun\_liu@sutd.edu.sg } \\
}

\begin{document}
\maketitle

\blfootnote{\dag~Equal contribution;~~\ddag~Corresponding author} 

\begin{abstract}
Estimating the 6D object pose from a single RGB image often involves noise and indeterminacy due to challenges such as occlusions and cluttered backgrounds. 
Meanwhile, diffusion models have shown appealing performance in generating high-quality images from random noise with high indeterminacy through step-by-step denoising. 
Inspired by their denoising capability, we propose a novel diffusion-based framework (\textbf{6D-Diff}) to handle the noise and indeterminacy in object pose estimation for better performance. 
In our framework, to establish accurate 2D-3D correspondence, we formulate 2D keypoints detection as a reverse diffusion (denoising) process.
To facilitate such a denoising process, we design a Mixture-of-Cauchy-based forward diffusion process and condition the reverse process on the object appearance features.
Extensive experiments on the LM-O and YCB-V datasets demonstrate the effectiveness of our framework.
\end{abstract}

\section{Introduction}
6D object pose estimation aims to estimate the 6D pose of an object including its location and orientation, 
which has a wide range of applications, such as augmented reality \cite{marchand2015pose,Rambach20186DoFOT}, robotic manipulation 
\cite{perez2016robot,busam2015stereo}, and automatic driving \cite{wu20196d}. 
Recently, various methods \cite{peng2019pvnet,xu2022rnnpose,su2022zebrapose,castro2023crt,kehl2017ssd,hu2020single,wang2021gdr,chen2020end,hodan2020epos} have been proposed to conduct RGB-based 6D object pose estimation since RGB images are easy to obtain.
Despite the increased efforts, a variety of challenges persist in RGB-based 6D object pose estimation, including occlusions, cluttered backgrounds, and changeable environments \cite{xiang2018posecnn,peng2019pvnet, di2021so, wang2021occlusion, mei2022spatial}. 
These challenges can introduce significant noise and indeterminacy into the pose estimation process, leading to error-prone predictions \cite{peng2019pvnet, di2021so, mei2022spatial}.

Meanwhile, diffusion models \cite{NEURIPS2020_DDPM,song2021denoising} have achieved appealing results in various generation tasks such as image synthesis \cite{NEURIPS2020_DDPM, dhariwal2021diffusion} and image editing \cite{meng2021sdedit}. 
Specifically, diffusion models are able to recover high-quality determinate samples (e.g., clean images) from a noisy and indeterminate input data distribution (e.g., random noise) via a step-by-step denoising process \cite{NEURIPS2020_DDPM, song2021denoising}.
Motivated by such a strong denoising capability \cite{NEURIPS2020_DDPM,gu2022stochastic,gong2023diffpose},
we aim to leverage diffusion models to handle the RGB-based 6D object pose estimation task,
since this task also involves tackling noise and indeterminacy. 
However, it can be difficult to directly use diffusion models to estimate the object pose, because diffusion models often start denoising from random Gaussian noise \cite{NEURIPS2020_DDPM,song2021denoising}. 
Meanwhile, in RGB-based 6D object pose estimation, the object pose is often extracted from an intermediate representation, such as keypoint heatmaps \cite{chen2020end}, pixel-wise voting vectors \cite{peng2019pvnet}, or object surface keypoint features \cite{castro2023crt}. 
Such an intermediate representation encodes useful distribution priors about the object pose.
Thus starting denoising from such an representation shall effectively assist the diffusion model in recovering accurate object poses.
To achieve this, we propose a novel diffusion-based object pose estimation framework (\textbf{6D-Diff}) that can exploit prior distribution knowledge from the intermediate representation for better performance. 

\begin{figure}[t]
\centering
\includegraphics[width=\columnwidth]{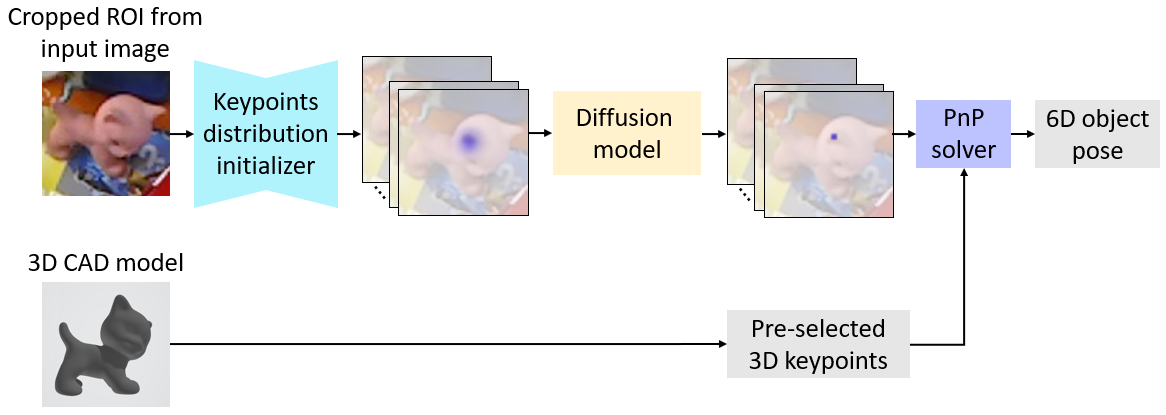}
\vspace{-0.3cm}
\caption{Overview of our proposed \textbf{6D-Diff} framework.
As shown, given the 3D keypoints from the object 3D CAD model, we aim to detect the corresponding 2D keypoints in the image to obtain the 6D object pose.
Note that when detecting keypoints, there are often challenges such as occlusions (including self-occlusions) and cluttered backgrounds that can introduce noise and indeterminacy into the process, impacting the accuracy of pose prediction.}
\vspace{-0.6cm}
\label{fig:intro}
\end{figure}

\begin{figure}[t]
\centering
\includegraphics[width=0.8\columnwidth]{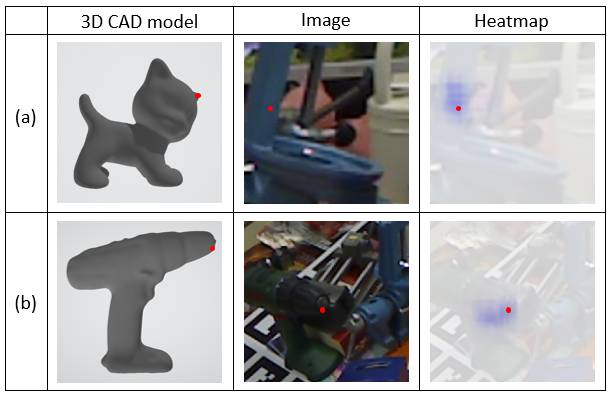}
\vspace{-0.4cm}
\caption{
Above we show two examples of keypoint heatmaps, which serve as the intermediate representation \cite{chen2020end, peng2019pvnet, castro2023crt} in our framework. 
The red dots indicate the ground-truth locations of the keypoints.
In the example (a), the target object is the pink cat, which is heavily occluded in the image and is shown in a different pose compared to the 3D model.
As shown above, due to occlusions and cluttered backgrounds, 
the keypoint heatmaps are noisy, which reflects the noise and indeterminacy during the keypoints detection process.
}
\vspace{-0.6cm}
\label{fig:demo}
\end{figure}

Overall, our framework is a \textit{correspondence-based} framework, in which to predict an object pose, given the 3D keypoints pre-selected from the object 3D CAD model, we first predict the coordinates of the 2D image keypoints corresponding to the pre-selected 3D keypoints. 
We then use the 3D keypoints together with the predicted 2D keypoints coordinates to compute the 6D object pose using a Perspective-n-Point (PnP) solver \cite{gao2003complete,lepetit2009ep}. 
As shown in Fig.~\ref{fig:intro}, to predict the 2D keypoints coordinates, we first extract an intermediate representation (the 2D keypoints heatmaps) through a keypoints distribution initializer. 
As discussed before, due to various factors, there often exists noise and indeterminacy in the keypoints detection process and the extracted heatmaps can be noisy as shown in Fig. \ref{fig:demo}.
Thus we pass the distribution modeled from these keypoints heatmaps into a diffusion model to perform the denoising process to obtain the final keypoints coordinates prediction.

Analogous to non-equilibrium thermodynamics \cite{sohl2015deep}, 
given a 2D image keypoint, we can consider all its possible locations in the image as particles in thermodynamics. 
Under low indeterminacy, the particles (possible locations) w.r.t. each 2D keypoint gather, and each keypoint can be determinately and accurately localized.
In contrast, under high indeterminacy, these particles can stochastically spread over the input image, and it is difficult to localize each keypoint. 
The process of converting particles from low indeterminacy to high indeterminacy is called the \textit{forward process} of the diffusion model. 
The goal of the diffusion model is to reverse the above forward process (through a \textit{reverse process}), i.e., converting the particles from high indeterminacy to low indeterminacy.
Here in our case, we aim to convert the indeterminate keypoints coordinates distribution modeled from the heatmaps into the determinate distribution. 
Below we briefly introduce the forward process and the reverse process in our diffusion model.

In the forward process, we aim to generate supervision signals that will be used to optimize the diffusion model during the reverse process. 
Specifically, given a set of pre-selected 3D keypoints, we first acquire ground-truth coordinates of their corresponding 2D keypoints using the ground-truth object pose. 
Then these determinate ground-truth 2D coordinates are gradually diffused towards the indeterminate distribution modeled from the intermediate representation, and the distributions generated along the way will be used as supervision signals. 
Note that, as the distribution modeled from the intermediate representation can be complex and irregular, it is difficult to characterize such a distribution via the Gaussian distribution.
This means that simply applying diffusion models in most existing generation works \cite{NEURIPS2020_DDPM, song2021denoising, dhariwal2021diffusion}, which start denoising from the random Gaussian noise, can introduce potentially large errors. 
To tackle this challenge, we draw inspiration from the fact that the Mixture of Cauchy (MoC) model can effectively characterize complex and intractable distributions. 
Moreover, the MoC model is robust to potential outliers in the distribution to be characterized \cite{sym11091186}.
Thus we propose to model the intermediate representation using a MoC distribution instead of simply treating it as a random Gaussian noise.
In this way, we gradually diffuse the determinate distribution (ground truth) of keypoints coordinates towards the modeled MoC distribution during the forward process. 

Correspondingly, in the reverse process, starting from the MoC distribution modeled in the forward process, we aim to learn to recover the ground-truth keypoints coordinates.
To achieve this, we leverage the distributions generated step-by-step during the forward process as the supervision signals to train the diffusion model to learn the reverse process. 
In this way, the diffusion model can learn to convert the indeterminate MoC distribution of keypoints coordinates into a determinate one smoothly and effectively. 
After the reverse process, the 2D keypoints coordinates obtained from the final determinate distribution are used to compute the 6D object pose with the pre-selected 3D keypoints.
Moreover, we further facilitate the model learning of such a reverse process by injecting object appearance features as context information.

Our work makes the following contributions.
1) We propose a novel \textbf{6D-Diff} framework, in which we formulate keypoints detection for 6D object pose estimation as a reverse diffusion process to effectively eliminate the noise and indeterminacy in object pose estimation.
2) To take advantage of the intermediate representation that encodes useful prior distribution knowledge for handling this task, we propose a novel MoC-based diffusion process. Besides, we facilitate the model learning by utilizing object features.

\section{Related Work}
\noindent\textbf{RGB-based 6D Object Pose Estimation} has received a lot of attention \cite{xiang2018posecnn,park2019pix2pose,peng2019pvnet,rad2017bb8,tekin2018real,su2022zebrapose,castro2023crt,Iwase_2021_ICCV,li2018deepim,Manhardt_2018_ECCV,Sundermeyer2018Implicit3O,Zakharov2019DPOD6P,xu2022rnnpose,haugaard2022surfemb,li2022dcl,liu2022gdrnpp_bop,yang2023object,guo2023knowledge,hai2023rigidity,hai2023shape, yang20236d}. 
Some works \cite{xiang2018posecnn,kehl2017ssd, hu2020single, wang2021gdr} proposed to directly regress object poses. 
However, the non-linearity of the rotation space makes direct regression of object poses difficult \cite{li2022dcl}. 
Compared to this type of \textit{direct methods}, \textit{correspondence-based methods} \cite{chen2020end,hodan2020epos,park2019pix2pose,peng2019pvnet,rad2017bb8,tekin2018real,su2022zebrapose} often demonstrate better performance, which estimate 6D object poses via learning 2D-3D correspondences between the observed image and the object 3D model.

Among \textit{correspondence-based methods}, several works \cite{rad2017bb8,peng2019pvnet,tekin2018real,oberweger2018making,ren2022robust} aim to predict the 2D keypoints coordinates corresponding to specific 3D keypoints. 
BB8 \cite{rad2017bb8} proposed to detect the 2D keypoints corresponding to the 8 corners of the object’s 3D bounding box. 
Later, PVNet \cite{peng2019pvnet} achieved better performance by estimating 2D keypoints for sampled points on the surface of the object 3D model via pixel-wise voting.
Moreover, various methods \cite{park2019pix2pose,Zakharov2019DPOD6P,hodan2020epos,wang2021gdr,su2022zebrapose} establish 2D-3D correspondences by localizing the 3D model point corresponding to each observed object pixel.
Among these methods, DPOD \cite{Zakharov2019DPOD6P} explored the use of UV texture maps to facilitate model training,
and ZebraPose \cite{su2022zebrapose} proposed to encode the surface of the object 3D model efficiently through a hierarchical binary grouping.
Besides, several pose refinement methods  \cite{Iwase_2021_ICCV,li2018deepim,Manhardt_2018_ECCV,xu2022rnnpose} have been proposed, which conducted pose refinement given an initial pose estimation.

In this paper, we also regard object pose estimation as a 2D-3D correspondence estimation problem. 
Different from previous works, here by formulating 2D-3D correspondence estimation as a distribution transformation process (denoising process), 
we propose a new framework (\textbf{6D-Diff}) that trains a diffusion model to perform progressive denoising from an indeterminate keypoints distribution to the desired keypoints distribution with low indeterminacy.

\noindent\textbf{Diffusion Models} \cite{NEURIPS2020_DDPM,song2021denoising,dhariwal2021diffusion, sohl2015deep, foo2023aigc} are originally introduced for image synthesis. 
Showing appealing generation capabilities, diffusion models have also been explored in various other tasks \cite{meng2021sdedit,gu2022stochastic,lugmayr2022repaint,gong2023diffpose, jiang2024se, urain2022se3dif,lee2023bias,hsiao2023confronting}, such as image editing \cite{meng2021sdedit} and image impainting \cite{lugmayr2022repaint}.
Here we explore a new framework that tackles object pose estimation with a diffusion model.
Different from previous generation works \cite{dhariwal2021diffusion, meng2021sdedit, lugmayr2022repaint} that start denoising from random noise, to aid the denoising process for 6D object pose estimation, we design a novel MoC-based diffusion mechanism that enables the diffusion model to start denoising from a distribution containing useful prior distribution knowledge regarding the object pose.
Moreover, we condition the denoising process on the object appearance features,
to further guide the diffusion model to obtain accurate predictions.

\section{Method}

To handle the noise and indeterminacy in RGB-based 6D object pose estimation, inspired by \cite{gong2023diffpose},
from a novel perspective of distribution transformation with progressive denoising,
we propose a framework (\textbf{6D-Diff}) that represents a new brand of diffusion-based solution for 6D object pose estimation.
Below we first revisit diffusion models in Sec.~\ref{Sec:revisiting}.  
Then we discuss our proposed framework in Sec.~\ref{Sec:training}, and introduce its training and testing scheme in Sec.~\ref{Sec:overall}. 
We finally detail the model architecture in Sec.~\ref{Sec:architecture}. 

\subsection{Revisiting Diffusion Models}
\label{Sec:revisiting}

The diffusion model \cite{NEURIPS2020_DDPM,song2021denoising}, which is a kind of probabilistic generative model, consists of two parts, namely the forward process and the reverse process.
Specifically, given an original sample $d_0$ (e.g., a clean image), the process of diffusing the sample $d_0$ iteratively towards the noise (typically Gaussian noise) $d_K \sim \mathcal{N}(\textbf{0}, \textbf{I})$ (i.e., $d_0 \rightarrow d_1 \rightarrow ... \rightarrow d_K$) is called the forward process. 
In contrast, the process of denoising the noise $d_K$ iteratively towards the sample $d_0$ (i.e., $d_K \rightarrow d_{K-1} \rightarrow ... \rightarrow d_0$) is called the reverse process.
Each process is defined as a Markov chain.

\noindent\textbf{Forward Process.} 
To obtain supervision signals for training the diffusion model to learn to perform the reverse process in a stepwise manner,
we need to acquire the intermediate step results $\{d_k\}^{K-1}_{k=1}$. 
Thus the forward process is first performed to generate these intermediate step results for training purpose.
Specifically, the posterior distribution $q(d_{1:K}|d_0)$ from $d_1$ to $d_K$ is formulated as:
\begin{equation} \label{eq:revisiting_1}
\setlength{\abovedisplayskip}{3pt}
\setlength{\belowdisplayskip}{3pt}
\begin{aligned}
& q(d_{1:K}|d_0) = \prod^K_{k=1}q(d_k|d_{k-1}) \\
& q(d_k|d_{k-1}) = \mathcal{N}(d_k; \sqrt{1-\beta_k}d_{k-1}, \beta_k\textbf{I})
\end{aligned}
\end{equation}
where $\{\beta_k \in (0, 1)\}^K_{k=1}$ denotes a set of fixed variance controllers that control the scale of the injected noise at different steps. 
According to Eq.~\eqref{eq:revisiting_1}, we can derive $q(d_k|d_0)$ in closed form as:
\begin{equation} \label{eq:revisiting_2}
\setlength{\abovedisplayskip}{3pt}
\setlength{\belowdisplayskip}{3pt}
\begin{aligned}
& q(d_k|d_0) = \mathcal{N}(d_k; \sqrt{\overline{\alpha}_k}d_0, (1-\overline{\alpha}_k)\textbf{I})
\end{aligned}
\end{equation}
where $\alpha_k = 1 - \beta_k$ and $\overline{\alpha}_k = \prod^k_{s=1} \alpha_s$.  
Based on Eq.~\eqref{eq:revisiting_2}, $d_k$ can be further expressed as:
\begin{equation} \label{eq:revisiting_3}
\setlength{\abovedisplayskip}{3pt}
\setlength{\belowdisplayskip}{3pt}
\begin{aligned}
& d_k = \sqrt{\overline{\alpha}_k}d_0 + \sqrt{1 - \overline{\alpha}_k}\epsilon
\end{aligned}
\end{equation}
where $\epsilon \sim \mathcal{N}(\textbf{0}, \textbf{I})$. 
From Eq.~\eqref{eq:revisiting_3}, we can observe that when the number of diffusion steps $K$ is sufficiently large and $\overline{\alpha}_K$ correspondingly decreases to nearly zero, 
the distribution of $d_K$ is approximately a standard Gaussian distribution, i.e.,  $d_K \sim \mathcal{N}(\textbf{0},\textbf{I})$. 
This means $d_0$ is gradually corrupted into Gaussian noise, which conforms to the non-equilibrium thermodynamics phenomenon of the diffusion process \cite{sohl2015deep}.

\noindent\textbf{Reverse Process.} 
With the intermediate step results $\{d_k\}^{K-1}_{k=1}$ acquired in the forward process, the diffusion model is trained to learn to perform the reverse process. 
Specifically, in the reverse process, each step can be formulated as a function $f$ that takes $d_k$ and the diffusion model $M_{diff}$ as inputs and generate $d_{k-1}$ as the output, i.e., $d_{k-1} = f(d_k, M_{diff})$.

After training the diffusion model, during inference, we do not need to conduct the forward process. Instead, we only conduct the reverse process, which converts a random Gaussian noise $d_K \sim \mathcal{N}(\textbf{0}, \textbf{I})$ into a sample $d_0$ of the desired distribution using the trained diffusion model.

\begin{figure*}[t]
\centering
\includegraphics[width=0.9\textwidth]{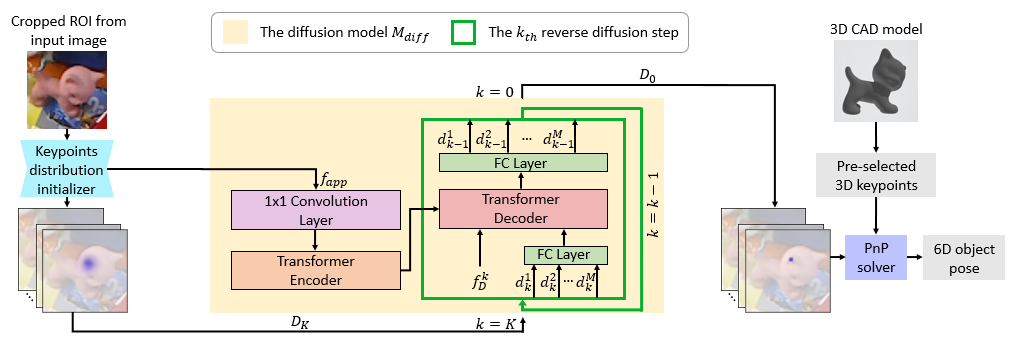}
\caption{Illustration of our framework. 
During testing, given an input image, we first crop the Region of Interest (ROI) from the image through an object detector. 
After that, we feed the cropped ROI to the keypoints distribution initializer to obtain the heatmaps that can provide useful distribution priors about keypoints, to initialize $D_K$. 
Meanwhile, we can obtain object appearance features $f_{\text{app}}$. 
Next, we pass $f_{\text{app}}$ into the encoder, and the output of the encoder will serve as conditional information to aid the reverse process in the decoder. 
We sample $M$ sets of 2D keypoints coordinates from $D_K$, 
and feed these $M$ sets of coordinates into the decoder to perform the reverse process iteratively together with the step embedding $f^k_D$.
At the final reverse step ($K$-th step), we average $\{d^i_0\}^{M}_{i=1}$ as the final keypoints coordinates prediction $d_0$, and use $d_0$ to compute the 6D pose with the pre-selected 3D keypoints via a PnP solver.}
\vspace{-0.5cm}
\label{fig:framework}
\end{figure*}

\subsection{Proposed Framework}
\label{Sec:training}
Similar to previous works \cite{su2022zebrapose, peng2019pvnet, hu2019segmentation}, our framework predicts 6D object poses via a two-stage pipeline.
Specifically, (i) we first select $N$ 3D keypoints on the object CAD model and detect the corresponding $N$ 2D keypoints in the image; 
(ii) we then compute the 6D pose using a PnP solver. 
Here we mainly focus on the first stage and aim to produce more accurate keypoint detection results. 

When detecting 2D keypoints, factors like occlusions and cluttered backgrounds can bring noise and indeterminacy into this process, and affect the accuracy of detection results \cite{peng2019pvnet, hu2019segmentation}.
To handle this problem, inspired by that diffusion models can iteratively reduce indeterminacy and noise in the initial distribution (e.g., standard Gaussian distribution) to generate determinate and high-quality samples of the desired distribution \cite{gu2022stochastic,gong2023diffpose}, 
we formulate keypoints detection as generating a determinate distribution of keypoints coordinates ($D_0$) from an indeterminate initial distribution ($D_K$) via a diffusion model.

Moreover, to effectively adapt to the 6D object pose estimation task, the diffusion model in our framework does not start the reverse process from the common initial distribution 
(i.e., the standard Gaussian distribution) as in most existing diffusion works \cite{NEURIPS2020_DDPM, dhariwal2021diffusion, song2021denoising}.
Instead, inspired by recent 6D object pose estimation works \cite{castro2023crt, wang2021gdr, chen2020end}, we first extract an intermediate representation (e.g., heatmaps), and use this representation to initialize a keypoints coordinates distribution (i.e., $D_K$), which will serve as the starting point of the reverse process.
Such an intermediate representation encodes useful prior distribution information about keypoints coordinates.
Thus by starting the reverse process from this representation, we effectively exploit the distribution priors in the representation to aid the diffusion model in recovering accurate keypoints coordinates. 
Below, we first describe how we initialize the keypoints distribution $D_K$, and then discuss the corresponding forward and reverse processes in our new framework.

\noindent\textbf{Keypoints Distribution Initialization.} 
We initialize the keypoints coordinates distribution $D_K$ with extracted heatmaps.
Specifically, similar to \cite{su2022zebrapose, li2019cdpn, labbe2020cosypose}, we first use an off-the-shelf object detector (e.g., Faster RCNN \cite{ren2015faster}) to detect the bounding box of the target object, and then crop the detected Region of Interest (ROI) from the input image.
We send the ROI into a sub-network (i.e., the keypoints distribution initializer) to predict a number of heatmaps where each heatmap corresponds to one 2D keypoint.
We then normalize each heatmap to convert it to a probability distribution.
In this way, each normalized heatmap naturally represents the distribution of the corresponding keypoint coordinates, and thus
we can use these heatmaps to initialize $D_K$.

\noindent\textbf{Forward Process.} 
After distribution initialization, the next step is to iteratively reduce the noise and indeterminacy in the initialized distribution $D_K$ by performing the reverse process ($D_K \rightarrow D_{K-1} \rightarrow ... \rightarrow D_0$). 
To train the diffusion model to perform such a reverse process, we need to obtain the distributions generated along the way
(i.e., $\{D_k\}^{K-1}_{k=1}$) as the supervision signals.
Thus, we first need to conduct the forward process to obtain samples from $\{D_k\}^{K-1}_{k=1}$.
Specifically, given the ground-truth keypoints coordinates distribution $D_0$, we define the forward process as: $D_0 \rightarrow D_1 \rightarrow ... \rightarrow D_K$, where $K$ is the number of diffusion steps.
In this forward process, we iteratively add noise to the determinate distribution $D_0$, i.e., increasing the indeterminacy of generated distributions, to transform it into the initialized distribution $D_K$ with indeterminacy.   
Via this process, we can generate $\{D_k\}^{K-1}_{k=1}$ along the way and use them as supervision signals to train the diffusion model to perform the reverse process.

However, in our framework, we do not aim to transform the ground-truth keypoints coordinates distribution $D_0$ towards a standard Gaussian distribution via the forward process, 
because our initialized distribution $D_K$ is not a random noise.
Instead, as discussed before, $D_K$ is initialized with heatmaps (as shown in Fig. \ref{fig:framework}), since the heatmaps can provide rough estimations about the keypoints coordinates distribution. 
To effectively utilize such priors in $D_K$ to facilitate the reverse process, we aim to enable the diffusion model to start the reverse process (denoising process) from $D_K$ instead of random Gaussian noise.  
Thus, the basic forward process (described in Sec. \ref{Sec:revisiting}) in existing generative diffusion models is not suitable in our framework, which motivates us to design a new forward process for our task.

\begin{figure*}[t]
\centering
\includegraphics[width=0.8\linewidth]{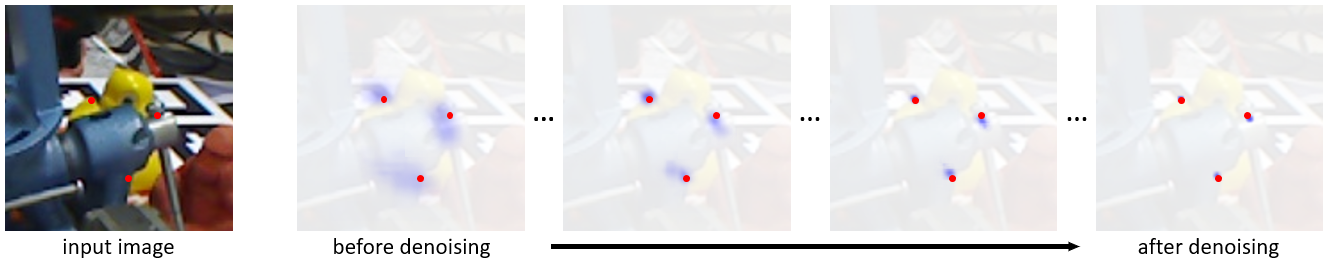}
\vspace{-0.3cm}
\caption{Visualization of the denoising process of a sample with our framework.
In this example, the target object is the yellow duck and for clarity, we here show three keypoints only.
The red dots indicate the ground-truth locations of these three keypoints.
The noisy heatmap before denoising reflects that factors like occlusions and clutter in the scene can introduce noise and indeterminacy when detecting keypoints.
As shown, our diffusion model can effectively and smoothly reduce the noise and indeterminacy in the initial distribution step by step, finally recovering a high-quality and determinate distribution of keypoints coordinates. (Better viewed in color)}
\vspace{-0.3cm}
\label{fig:denoise}
\end{figure*}

However, it is non-trivial to design such a forward process, as the initialized distribution $D_K$ is based on extracted heatmaps, 
and thus $D_K$ can be complex and irregular, as shown in Fig. \ref{fig:denoise}.
Hence modeling $D_K$ as a Gaussian distribution can result in potentially large errors.
To handle this challenge, motivated by that the Mixture of Cauchy (MoC) model can effectively and reliably characterize complex and intractable distributions \cite{sym11091186}, we leverage MoC to characterize $D_K$.
Based on the characterized distribution, we can then perform a corresponding MoC-based forward process.

Specifically, we denote the number of Cauchy kernels in the MoC distribution as $U$, and use the Expectation-Maximum-type (EM) algorithm \cite{sym11091186,teimouri2018statistical} to optimize the MoC parameters $\eta^{\text{MoC}}$ to characterize the distribution $D_K$ as:
\begin{equation} \label{eq:training_1}
\setlength{\abovedisplayskip}{3pt}
\setlength{\belowdisplayskip}{3pt}
\begin{aligned}
\eta_{*}^{\text{MoC}} = \text{EM} \Big( \prod^V_{v=1} \sum^U_{u=1} \pi_u \text{Cauchy}(d^v_K|\mu_u, \gamma_u) \Big)
\end{aligned}
\end{equation}
where $\{d^v_K\}^{V}_{v=1}$ denotes $V$ sets of keypoints coordinates sampled from the distribution $D_K$.
Note each set of keypoints coordinates $d^v_K$ contains all the $N$ keypoints coordinates (i.e., $ d^v_K \in \mathbb{R}^{N\times2}$).
$\pi_u$ denotes the weight of the $u$-th Cauchy kernel ($\sum^U_{u=1} \pi_u$ = 1), and $\eta^{\text{MoC}} = \{\mu_1, \gamma_1, ..., \mu_U, \gamma_U\}$ denotes the MoC parameters in which $\mu_u$ and $\gamma_u$ are the location and scale of the $u$-th Cauchy kernel. 
Via the above optimization, we can use the optimized parameters $\eta_{*}^{\text{MoC}}$ to model $D_K$ as the characterized distribution ($\hat{D}_K$). 
Given $\hat{D}_K$, we aim to conduct the forward process from the ground-truth keypoints coordinates distribution $D_0$, so that after $K$ steps of forward diffusion, the generated distribution reaches $\hat{D}_K$.
To this end, we modify Eq.~\eqref{eq:revisiting_3} as follows:
\begin{equation} \label{eq:training_2}
\begin{aligned}
& \hat{d}_k = \sqrt{\overline{\alpha}_k}d_0 + (1 - \sqrt{\overline{\alpha}_k}) \mu^{\text{MoC}} +
\sqrt{1 - \overline{\alpha}_k}\epsilon^{\text{MoC}}
\end{aligned}
\end{equation}
where $\hat{d}_k\in\mathbb{R}^{N\times2}$ represents a sample (i.e., a set of $N$ keypoints coordinates) from the generated distribution $\hat{D}_k$, 
$\mu^{\text{MoC}}=\sum^U_{u=1} \mathds{1}_u \mu_u$, and $\epsilon^{\text{MoC}} \sim \text{Cauchy}(\textbf{0}, \sum^U_{u=1} (\mathds{1}_u \gamma_u))$. Note that $\mathds{1}_u$ is a zero-one indicator and $\sum^U_{u=1} \mathds{1}_u = 1$ and $\text{Prob}(\mathds{1}_u = 1) = \pi_u$.

From Eq.~\eqref{eq:training_2}, we can observe that when $K$ is sufficiently large and $\overline{\alpha}_K$ correspondingly decreases to nearly zero, the distribution of $\hat{d}_K$ reaches the MoC distribution, i.e., $\hat{d}_K = \mu^{\text{MoC}} + \epsilon^{\text{MoC}} \sim \text{Cauchy}(\sum^U_{u=1} (\mathds{1}_u \mu_u), \sum^U_{u=1} (\mathds{1}_u \gamma_u))$. 
After the above MoC-based forward process, we can use the generated 
$\{\hat{D}_k\}^{K-1}_{k=1}$ as supervision signals
to train the diffusion model $M_{\text{diff}}$ to learn the reverse process.
More details about Eq.~\eqref{eq:training_2} can be found in Supplementary material.
Such a forward process is only conducted to generate supervision signals for training the diffusion model, while we only need to conduct the reverse process during testing.

\noindent\textbf{Reverse Process.} 
In the reverse process, we aim to recover a desired determinate keypoints distribution $D_0$ from the 
initial distribution $D_K$.
As discussed above, we characterize $D_K$ via a MoC model and then generate 
$\{\hat{D}_k\}^{K-1}_{k=1}$ 
as supervision signals
to optimize the diffusion model to learn to perform the reverse process ($\hat{D}_K \to \hat{D}_{K-1} \to ... \to D_0$), in which the model iteratively reduces the noise and indeterminacy in $\hat{D}_K$ to generate $D_0$.

However, it can still be difficult to generate $D_0$ by directly performing the reverse process from $\hat{D}_K$,
because the object appearance features are lacking in $\hat{D}_K$.
Such features can help constrain the model reverse process based on the input image to get accurate predictions. 
Thus we further leverage the appearance features from the image as context to guide $M_{\text{diff}}$ in the reverse process.
Specifically, we reuse the features extracted from the keypoints distribution initializer as the appearance features $f_{\text{app}}$ and feed $f_{\text{app}}$ into the diffusion model, as shown in Fig. \ref{fig:framework}.

Our reverse process aims to generate a determinate distribution $D_0$ from the indeterminate distribution $\hat{D}_K$ (during training) or $D_K$ (during testing). 
Below we describe the reverse process during testing. 
We first obtain $f_{\text{app}}$ from the input image.
Then to help the diffusion model to learn to perform denoising at each reverse step, following \cite{NEURIPS2020_DDPM, song2021denoising}, we generate the unique step embedding $f^k_{D}$ to inject the step number ($k$) information into the model.
In this way, given a set of noisy keypoints coordinates $d_k \in \mathbb{R}^{N\times2}$ drawn from ${D}_k$ at the $k^{th}$ step, we use diffusion model $M_{\text{diff}}$, conditioned on the step embedding $f^k_D$ and the object appearance features $f_{\text{app}}$, to recover ${d}_{k-1}$ from ${d}_{k}$ as:
\begin{equation} \label{eq:denoising}
\setlength{\abovedisplayskip}{3pt}
\setlength{\belowdisplayskip}{3pt}
{d}_{k-1} = M_{\text{diff}}({d}_k, f_{\text{app}}, f^k_D)
\end{equation}

\subsection{Training and Testing}
\label{Sec:overall}
\noindent\textbf{Training.} 
Following \cite{peng2019pvnet}, we first select $N$ 3D keypoints from the surface of the object CAD model using the farthest point sampling (FPS) algorithm. Then we conduct the training process in the following two stages. 

In the first stage, to initialize the distribution $D_K$, we optimize the keypoints distribution initializer.
Specifically, for each training sample, given the pre-selected $N$ 3D keypoints, we can obtain the ground-truth coordinates of the corresponding $N$ 2D keypoints using the ground-truth 6D object pose.
Then for each keypoint, based on the corresponding ground-truth coordinates, we generate a ground-truth heatmap following \cite{oberweger2018making} for training the initializer.
Thus for each training sample, we generate $N$ ground-truth heatmaps.
In this way, the loss function $L_{\text{init}}$ for optimizing the initializer can be formulated as:
\begin{equation}\label{eq:loss1}
\setlength{\abovedisplayskip}{3pt}
\setlength{\belowdisplayskip}{3pt}
\begin{aligned}
L_{\text{init}} = {\Big\Vert \textbf{H}_{\text{pred}} - \textbf{H}_{\text{GT}} \Big\Vert}^2_2
\end{aligned}
\end{equation}
where $\textbf{H}_{\text{pred}}$ and $\textbf{H}_{\text{GT}}$ denote the predicted heatmaps and ground-truth heatmaps, respectively.

In the second stage, we optimize the diffusion model $M_{\text{diff}}$. 
For each training sample, to optimize $M_{\text{diff}}$, we perform the following steps. 
(1) We first send the input image into an off-the-shelf object detector \cite{tian2019fcos} and then feed the detected ROI into the trained initializer to obtain $N$ heatmaps.
Meanwhile, we can also obtain $f_{\text{app}}$. 
(2) We use the $N$ predicted heatmaps to initialize $D_K$, and leverage the EM-type algorithm to characterize $D_K$ as a MoC distribution $\hat{D}_K$. 
(3) Based on $\hat{D}_K$, we use the ground-truth keypoints coordinates $d_0$ to directly generate $M$ sets of ($\hat{d}_1,..., \hat{d}_K$) (i.e., $\{\hat{d}^i_1,..., \hat{d}^i_K\}^{M}_{i=1}$) via the forward process (Eq. \eqref{eq:training_2}).
(4) Then, we aim to optimize the diffusion model $M_{\text{diff}}$ to recover $\hat{d}^i_{k-1}$ from $\hat{d}^i_k$ iteratively. Following previous diffusion works \cite{NEURIPS2020_DDPM, song2021denoising}, we formulate the loss $L_{\text{diff}}$ for optimizing $M_{\text{diff}}$ as follows ($\hat{d}^i_0=d_0$ for all $i$):
\begin{equation}\label{eq:loss1}
\setlength{\abovedisplayskip}{3pt}
\setlength{\belowdisplayskip}{3pt}
\begin{aligned}
L_{\text{diff}} = \sum^M_{i=1}\sum^K_{k=1} {\Big\Vert M_{\text{diff}}(\hat{d}^i_{k}, f_{\text{app}}, f^k_D) - \hat{d}^i_{k-1} \Big\Vert}^2_2
\end{aligned}
\end{equation}

\noindent\textbf{Testing.} 
During testing, for each testing sample, by feeding the input image to the object detector and the keypoints distribution initializer consecutively, we can initialize $D_K$ and meanwhile obtain $f_{\text{app}}$. 
Then, we perform the reverse process.
During the reverse process, we sample $M$ sets of noisy keypoints coordinates from $D_K$ (i.e., $\{d^i_K\}^{M}_{i=1}$) and feed them into the trained diffusion model.
Here we sample $M$ sets of keypoints coordinates, because we are converting from a distribution ($D_K$) towards another distribution ($D_0$).
Then the model iteratively performs the reverse steps.
After $K$ reverse diffusion steps, we obtain $M$ sets of predicted keypoints coordinates (i.e., $\{d^i_0\}^{M}_{i=1}$).
To obtain the final keypoints coordinates prediction $d_0$, we compute the mean of the $M$ predictions. 
Finally, we can solve for the 6D object pose using a PnP solver, like \cite{peng2019pvnet,su2022zebrapose}.

\subsection{Model Architecture}
\label{Sec:architecture}
Our framework mainly consists of the diffusion model ($M_{\text{diff}}$) and the keypoints distribution initializer.

\noindent\textbf{Diffusion Model} $M_{\text{diff}}$.
As illustrated in Fig.~\ref{fig:framework}, our proposed diffusion model $M_{\text{diff}}$ mainly consists of a transformer encoder-decoder architecture.
The appearance features $f_{\text{app}}$ 
are sent into the encoder for extracting context information to aid the reverse process in the decoder. 
$f^k_D$ and $\{d^i_k\}^{M}_{i=1}$ (or $\{\hat{d}^i_k\}^{M}_{i=1}$ during training) are sent into the decoder for the reverse process.
Both the encoder and the decoder contain a stack of three transformer layers.

\begin{table*}
\caption{Comparisons with RGB-based 6D object pose estimation methods on the LM-O dataset. (*) denotes symmetric objects.}
  \vspace{-0.3cm}
  \centering
  \resizebox{0.93\linewidth}{!}{
  \begin{tabular}{@{}c|c|c|c|c|c|c|c|c|c|c@{}}
    \hline
     Method & PVNet \cite{peng2019pvnet} & HybridPose~\cite{song2020hybridpose} & RePose~\cite{iwase2021repose} & DeepIM \cite{li2018deepim} &  GDR-Net~\cite{wang2021gdr} &  SO-Pose~\cite{di2021so} & CRT-6D~\cite{castro2023crt} & ZebraPose~\cite{su2022zebrapose} & CheckerPose~\cite{Lian_2023_ICCV} & \textbf{~~~Ours~~~} \\
     \hline
     ape & 15.8 & 20.9 & 31.1 & 59.2 & 46.8 & 48.4 & 53.4 & 57.9 & 58.3 & \textbf{60.6}\\
     can & 63.3 & 75.3 & 80.0 & 63.5 & 90.8 & 85.8 & 92.0 & 95.0 & 95.7 & \textbf{97.9}\\
     cat & 16.7 & 24.9 & 25.6 & 26.2 & 40.5 & 32.7 & 42.0 & 60.6 & 62.3 & \textbf{63.2}\\
     driller & 65.7 & 70.2 & 73.1 & 55.6 & 82.6 & 77.4 & 81.4 & 94.8 & 93.7 & \textbf{96.6}\\
     duck & 25.2 & 27.9 & 43.0 & 52.4 & 46.9 & 48.9 & 44.9 & 64.5 & \textbf{69.9} & 67.2\\
     eggbox* & 50.2 & 52.4 & 51.7 & 63.0 & 54.2 & 52.4 & 62.7 & 70.9 & 70.0 & \textbf{73.5}\\
     glue* & 49.6 & 53.8 & 54.3 & 71.7 & 75.8 & 78.3 & 80.2 & 88.7 & 86.4 & \textbf{92.0}\\
     holepuncher & 39.7 & 54.2 & 53.6 & 52.5 & 60.1 & 75.3 & 74.3 & 83.0 & 83.8 & \textbf{85.5}\\
     \hline
     Mean & 40.8 & 47.5 & 51.6 & 55.5 & 62.2 & 62.3 & 66.3 & 76.9 & 77.5 & \textbf{79.6}\\
     \hline
  \end{tabular}}
  \vspace{-0.4cm}
  \label{tab:lmo_results_table_}
\end{table*}
\begin{table}
\caption{Comparisons with RGB-based 6D object pose estimation methods on the YCB-V dataset. (-) indicates the corresponding result is not reported in the original paper.}
  \vspace{-0.3cm}
  \centering
  \resizebox{0.9\columnwidth}{!}{
  \begin{tabular}{@{}l|c|c|c@{}}
    \hline
     Method & ADD(-S) & AUC of ADD-S & AUC of ADD(-S) \\
    \hline
    SegDriven\cite{hu2019segmentation} & 39.0 &  - &  -  \\
    SingleStage\cite{hu2020single} & 53.9 &  - &  - \\
    CosyPose~\cite{labbe2020cosypose} & - &  89.8 &  84.5 \\
    RePose~\cite{iwase2021repose} & 62.1 &  88.5 &  82.0 \\
    GDR-Net~\cite{wang2021gdr} & 60.1 &  \textbf{91.6} &  84.4 \\
    SO-Pose~\cite{di2021so} & 56.8 &  90.9 &  83.9 \\
    ZebraPose~\cite{su2022zebrapose} & 80.5 &  90.1 &  85.3  \\
    CheckerPose~\cite{Lian_2023_ICCV} & 81.4 & 91.3 & 86.4 \\ \hline
    Ours & \textbf{83.8} & 91.5 & \textbf{87.0} \\
    \hline
  \end{tabular}}
\vspace{-0.4cm}
  \label{tab:ycbv_results_table}
\end{table}
More specifically, as for the encoder part, we first map $f_{\text{app}} \in \mathbb{R}^{16 \times 16 \times 512}$ through a 1 × 1 convolution layer to a latent embedding $e_{\text{app}} \in \mathbb{R}^{ 16 \times 16 \times 128 }$.
To retain the spatial information, following \cite{vaswani2017attention}, we further incorporate positional encodings into $e_{\text{app}}$.
Afterwards, we flatten $e_{\text{app}}$ into a feature sequence ($\mathbb{R}^{256 \times 128}$), and send it into the encoder.
The encoder output $f_{\text{enc}}$ containing the extracted object information will be sent into the decoder to aid the reverse process.
Note that during testing, for each sample, we only need to conduct the above computation process once to obtain the corresponding $f_{\text{enc}}$.

The decoder part iteratively performs the reverse process.
For notation simplicity, below we describe the reverse process for a single sample $d_k$ instead of the $M$ samples ($\{{d}^i_1,..., {d}^i_K\}^{M}_{i=1}$).
Specifically, at the $k$-th reverse step, to inject the current step number ($k$) information into the decoder, we first generate the step embedding $f^k_D \in \mathbb{R}^{1 \times 128}$ using the sinusoidal function following \cite{NEURIPS2020_DDPM, song2021denoising}.
Meanwhile, we use an FC layer to map the input $d_k \in \mathbb{R}^{N \times 2}$ to a latent embedding $e_{k} \in \mathbb{R}^{N \times 128}$.
Then we concatenate $f^k_D$ and $e_{k}$ along the first dimension, and send it into the decoder.
By interacting with the encoder output $f_{\text{enc}}$ (extracted object information) via cross-attention at each layer,
the decoder produces $f_{\text{dec}}$, which is further mapped into the keypoints coordinates prediction $d_{k-1} \in \mathbb{R}^{N \times 2}$ via an FC layer. 
Then we send $d_{k-1}$ back to the decoder as the input to perform the next reverse step.

\noindent\textbf{Keypoints Distribution Initializer.} The initializer adopts a ResNet-34 backbone, which is commonly used in 6D pose estimation methods \cite{wang2021gdr,su2022zebrapose,castro2023crt}. 
To generate heatmaps to initialize the distribution $D_K$, we add two deconvolution layers followed by a 1 × 1 convolution layer after the ResNet-34 backbone, and then we obtain predicted heatmaps $\textbf{H}_{\text{pred}} \in \mathbb{R}^{N \times \frac{H}{4} \times \frac{W}{4}}$ where $H$ and $W$ denote the height and width of the input ROI image respectively.
Moreover, the features outputted by the ResNet-34 backbone, combined with features obtained from methods \cite{su2022zebrapose, Lian_2023_ICCV}, are used as the object features $f_{\text{app}}$.

\section{Experiments}
\subsection{Datasets \& Evaluation Metrics}
Given that previous works \cite{di2021so, Zakharov2019DPOD6P, iwase2021repose} have reported the evaluation accuracy over 95\%  on the Linemod (LM) dataset \cite{hinterstoisser2013model}, the performance on this dataset has become saturated.
Thus recent works \cite{su2022zebrapose,castro2023crt} mainly focus on using the LM-O dataset \cite{brachmann2016uncertainty} and the YCB-V dataset \cite{xiang2018posecnn} that are more challenging, which we follow.

\noindent\textbf{LM-O Dataset.} 
The Linemod Occlusion (LM-O) dataset contains 1214 images and is a challenging subset of the LM dataset. 
In this dataset, around 8 objects are annotated on each image and the objects are often heavily occluded.
Following \cite{su2022zebrapose, castro2023crt}, we use both the real images from the LM dataset and the publicly available physically-based rendering (pbr) images \cite{denninger2019blenderproc} as the training images for LM-O. Following \cite{wang2021gdr,su2022zebrapose}, on LM-O dataset, we evaluate the model performance using the commonly-used ADD(-S) metric.
For this metric, we compute the mean distance between the model points transformed using the predicted pose and the same model points transformed using the ground-truth pose. For symmetric objects, following \cite{xiang2018posecnn}, the mean distance is computed based on the closest point distance.
If the mean distance is less than 10\% of the model diameter, the predicted pose is regarded as correct.

\noindent\textbf{YCB-V Dataset.} The YCB-V dataset is a large-scale dataset containing 21 objects and over 100k real images. 
The samples in this dataset often exhibit occlusions and cluttered backgrounds.
Following \cite{su2022zebrapose,castro2023crt}, we use both the real images from the training set of the YCB-V dataset and the publicly available pbr images as the training images for YCB-V. Following \cite{wang2021gdr,su2022zebrapose}, we evaluate the model performance using the following metrics: ADD(-S), AUC (Area Under the Curve) of ADD-S, and AUC of ADD(-S). 
For calculating AUC, we set the maximum distance threshold to 10 cm following \cite{xiang2018posecnn}.

\subsection{Implementation Details}

We conduct our experiments on an Nvidia V100 GPU.
We set the number of pre-selected 3D keypoints $N$ to 128.
During training, following \cite{su2022zebrapose, li2019cdpn}, we utilize the dynamic zoom-in strategy to produce augmented ROI images.
During testing, we use the detected bounding box with Faster RCNN \cite{ren2015faster} and FCOS \cite{tian2019fcos} provided by CDPNv2 \cite{li2019cdpn}. 
The cropped ROI image is resized to the shape of $3 \times 256 \times 256 $ ($H=W=256$).
We characterize $D_K$ via a MoC model with 9 Cauchy kernels ($U=9$) for the forward diffusion process.
We optimize the diffusion model $M_{\text{diff}}$ for 1500 epochs using the Adam optimizer \cite{kingma2014adam} with an initial learning rate of 4e-5.
Moreover, we set the number of sampled sets $M$ to 5, and the number of diffusion steps $K$ to 100. 
Following \cite{su2022zebrapose}, we use Progressive-X \cite{barath2019progressive} as the PnP solver.
Note that during testing, instead of performing the reverse process with all the $K$ steps, we accelerate the process with DDIM \cite{song2021denoising}, a recently proposed diffusion acceleration method. With DDIM acceleration, we only need to perform 10 steps to finish the reverse process during testing.

\subsection{Comparison with State-of-the-art Methods}

\begin{figure}[t]
\centering
\includegraphics[width=\columnwidth]{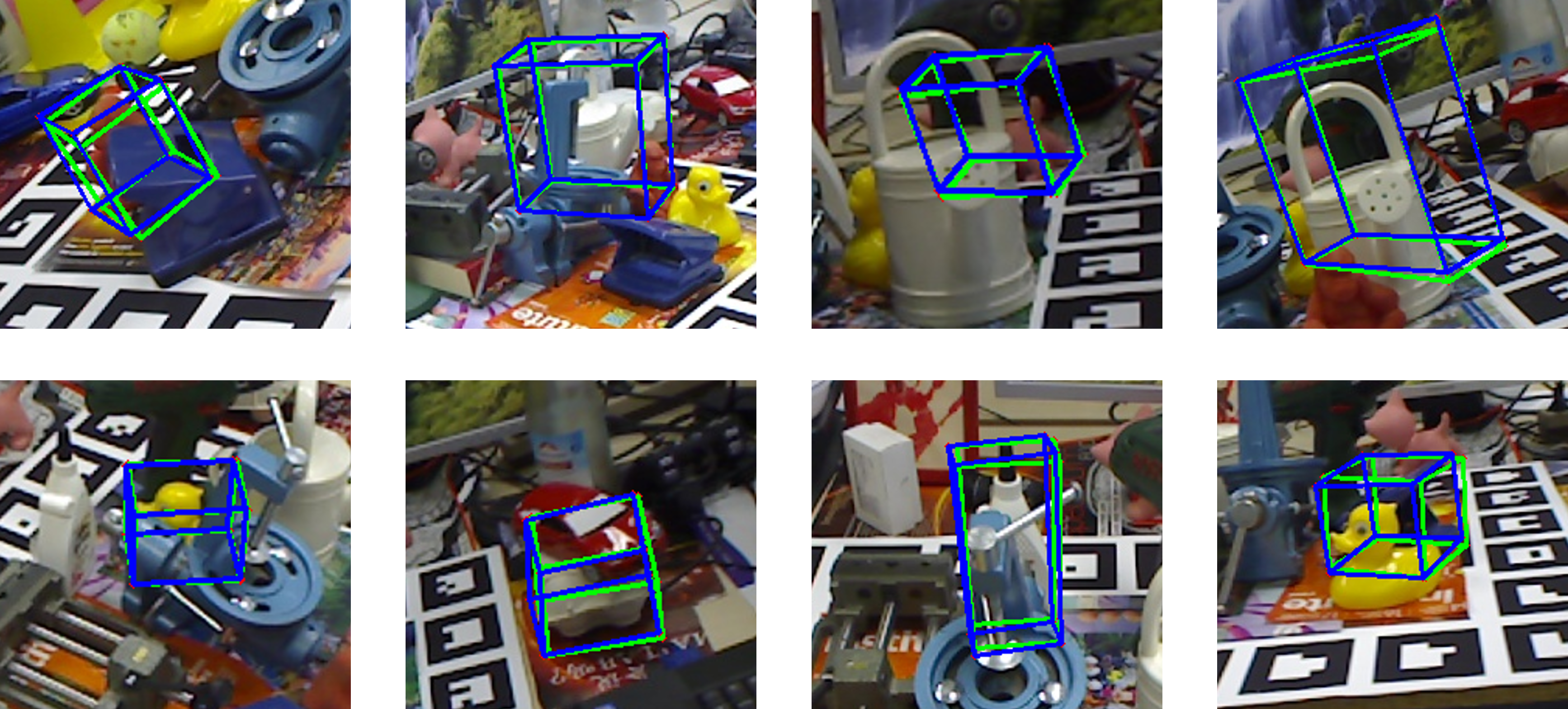}
\vspace{-0.3cm}
\caption{\textbf{Qualitative results}. 
\textcolor{green}{Green} bounding boxes represent the ground-truth poses and \textcolor{blue}{blue}
bounding boxes represent the predicted poses of our method. 
As shown, even facing severe occlusions, clutter in the scene or varying environment, our framework can still accurately recover the object poses, showing the effectiveness of our method for handling the noise and indeterminacy caused by various factors in object pose estimation.}
\vspace{-0.5cm}
\label{fig:visualization}
\end{figure}

\noindent\textbf{Results on LM-O Dataset.} 
As shown in Tab.~\ref{tab:lmo_results_table_},
compared to existing methods, our method achieves the best mean performance, showing the superiority of our method.
We also show qualitative results on the LM-O dataset in Fig. \ref{fig:visualization}. 
As shown, even in the presence of large occlusions (including self-occlusions) and cluttered backgrounds, 
our method still produces accurate predictions.

\noindent\textbf{Results on YCB-V Dataset.} 
As shown in Tab.~\ref{tab:ycbv_results_table}, our framework achieves the best performance on both the ADD(-S) and the AUC of ADD(-S) metrics, and is comparable to the state-of-the-art method on the AUC of ADD-S metric, showing the effectiveness of our method.

\subsection{Ablation Studies}

We conduct extensive ablation experiments on the LM-O dataset, and we report the model performance on ADD(-S) metric averaged over all the objects. 
\setlength{\columnsep}{0.15in}

\begin{wraptable}[8]{r}{0.4\columnwidth}
\vspace{-0.4cm}
\caption{Evaluation on the effectiveness of the denoising process.}
\vspace{-0.35cm}
\resizebox{0.4\columnwidth}{!}
{
\small
\begin{tabular}{l|c}
\hline
Method & ADD(-S)\\
\hline
Variant A & 49.2\\
Variant B & 57.3\\
Variant C & 61.1\\
\hline
6D-Diff & 79.6\\
\hline
\end{tabular}}
\label{Tab:ablation_study_1}
\end{wraptable}

\noindent\textbf{Impact of denoising process.} 
In our framework, we predict keypoints coordinates via performing the denoising process.
To evaluate the efficacy of this process, we test three variants.  
In the first variant (\textit{Variant A}), we remove the diffusion model $M_{\text{diff}}$ and predict keypoints coordinates directly from the heatmaps produced by the keypoints distribution initializer.
The second variant (\textit{Variant B}) has the same model architecture as our framework, but the diffusion model is optimized to directly predict the coordinates instead of learning the reverse process.
Same as \textit{Variant B}, the third variant (\textit{Variant C}) is also optimized to directly predict coordinates without denoising process.
For \textit{Variant C}, we stack our diffusion model structure multiple times to produce a deep network, which has similar computation complexity with our framework.
As shown in Tab.~\ref{Tab:ablation_study_1}, compared to our framework, the performance of these variants significantly drops, showing that the effectiveness of our framework mainly lies in the designed denoising process.
\begin{wraptable}[6]{r}{0.4\columnwidth}
\vspace{-0.35cm}
\caption{Evaluation on the effectiveness of the object appearance features $f_{\text{app}}$.}
\vspace{-0.35cm}
\resizebox{0.35\columnwidth}{!}
{
\small
\begin{tabular}{l|c}
\hline
Method & ADD(-S)\\
\hline
w/o $f_{\text{app}}$ & 74.4\\
\hline
6D-Diff & 79.6 \\
\hline
\end{tabular}}
\label{Tab:ablation_study_4}
\end{wraptable}

\noindent\textbf{Impact of object appearance features} $f_{\text{app}}$.
In our framework,  we send the appearance features $f_{\text{app}}$ 
into the diffusion model $M_{\text{diff}}$ to aid the reverse process.
To evaluate its effect, we test a variant in which we do not send $f_{\text{app}}$ into $M_{\text{diff}}$ (\textit{w/o $f_{\text{app}}$}).
As shown in Tab.~\ref{Tab:ablation_study_4}, our framework performs better than this variant, 
showing that $f_{\text{app}}$ can aid $M_{\text{diff}}$ to get more accurate predictions.
\begin{wraptable}[6]{r}{0.6\columnwidth}
\vspace{-0.4cm}
\caption{Evaluation on the effectiveness of the MoC design.}
\vspace{-0.35cm}
\resizebox{0.6\columnwidth}{!}
{
\small
\begin{tabular}{l|c}
\hline
Method & ADD(-S)\\
\hline
Standard diffusion w/o MoC & 73.1\\
Heatmaps as condition & 76.2\\
\hline
6D-Diff & 79.6\\
\hline
\end{tabular}}
\label{Tab:ablation_study_2}
\end{wraptable}

\noindent\textbf{Impact of MoC design.} 
During training, we model the distribution $D_K$ from the intermediate representation (heatmaps) as a MoC distribution $\hat{D}_K$, and train the diffusion model $M_{\text{diff}}$ to perform the reverse process from $\hat{D}_K$. 
To investigate the impact of this design, we evaluate two variants that train $M_{\text{diff}}$ in different ways.
In the first variant (\textit{Standard diffusion w/o MoC}), 
we train the model to start the reverse process from the standard Gaussian noise, i.e., following the basic forward process in Eq. \eqref{eq:revisiting_3} for model training.
In the second variant (\textit{Heatmaps as condition}), we still train the model to start denoising from the random Gaussian noise but we use the heatmaps as the condition for the reverse process.
As shown in Tab.~\ref{Tab:ablation_study_2}, our framework consistently outperforms both variants, showing effectiveness of the designed MoC-based forward process.

\section{Conclusion}
In this paper, we proposed a novel diffusion-based 6D object pose estimation framework, which effectively handles noise and indeterminacy in object pose estimation. 
In our framework, we formulate object keypoints detection as a carefully-designed reverse diffusion process. 
We design a novel MoC-based forward process to effectively utilize the distribution priors in intermediate representations.  
Our framework achieves superior performance. 

\noindent\textbf{Acknowledgement.} This work was supported by the National Research Foundation Singapore under the AI Singapore Programme (Award Number: AISG-100E-2023-121).

{
\small
\bibliographystyle{ieeenat_fullname}
\bibliography{main}
}


\end{document}